\documentclass[conference]{IEEEtran}
\IEEEoverridecommandlockouts
\usepackage{cite}
\usepackage{amsmath,amssymb,amsfonts}
\usepackage{graphicx}
\usepackage{textcomp}
\usepackage{xcolor}
\usepackage{amsmath,bm}
\usepackage{amsfonts}
\usepackage{amsthm}
\usepackage{graphicx}
\usepackage{cleveref}
\usepackage{algorithm}
\usepackage{algpseudocode}
\usepackage{caption}
\usepackage{subcaption}
\usepackage{textcomp}

\renewcommand{\baselinestretch}{0.97}

\usepackage{amsmath}

\newcommand{\vect}[1]{\boldsymbol{#1}}
\newcommand{\mat}[1]{\boldsymbol{#1}}

\newcommand{\diffs}[3]{\frac{\partial^2 #1}{
\ifx#2#3 
\partial #2^2
\else
\partial #2 \partial #3
\fi
}}

\newcommand{\av}{\vect{a}}

\newcommand{\pv}{\vect{p}}
\newcommand{\dpv}{\dot{\vect{p}}}

\newcommand{\qv}{{\vect{q}}}
\newcommand{\dqv}{\dot{\vect{q}}}

\newcommand{\xv}{\vect{x}}
\newcommand{\dxv}{\dot{\vect{x}}}

\newcommand{\IIm}{\mat{I}}

\newcommand{\Am}{\mat{A}}
\newcommand{\Bm}{\mat{B}}

\newcommand{\Jm}{\mat{J}}

\newcommand{\Km}{\mat{K}}

\newcommand{\Pm}{\mat{P}}

\newcommand{\Vm}{\mat{V}}

\makeatletter
\algrenewcommand\ALG@beginalgorithmic{\footnotesize}
\makeatother

\def\BibTeX{{\rm B\kern-.05em{\sc i\kern-.025em b}\kern-.08em
    T\kern-.1667em\lower.7ex\hbox{E}\kern-.125emX}}
\begin{document}

\title{Motion Control of Redundant Robots\\with Generalised Inequality Constraints
}

\author{Amirhossein Kazemipour$^\ast$ \qquad Maram Khatib$^\ast$ \qquad Khaled Al Khudir$^{\ast \ast}$ \qquad Alessandro De Luca$^\ast$
\thanks{$^\ast$Dipartimento di Ingegneria Informatica, Automatica e Gestionale, Sapienza Universit\`a di Roma, Via Ariosto 25, 00185 Roma, Italy. Emails: amrkzp@gmail.com, khatib@diag.uniroma1.it, deluca@diag.uniroma1.it.}
\thanks{$^{\ast \ast}$School of Mechanical, Aerospace
and Automotive Engineering, Coventry University, CV1 5FB Coventry, UK. Email: khaled.alkhudir@coventry.ac.uk.}}

\maketitle

\begin{abstract}
We present an improved version of the Saturation in the Null Space (SNS) algorithm for redundancy resolution at the velocity level. In addition to hard bounds on joint space motion, we consider also Cartesian box constraints that cannot be violated at any time. The modified algorithm combines all bounds into a single augmented generalised vector and gives equal, highest priority to all inequality constraints. When needed, feasibility of the original task is enforced by the SNS task scaling procedure. Simulation results are reported for a 6R planar robot.
\end{abstract}

\begin{IEEEkeywords}
Motion control, Redundant robots, Inequality constraints, Hard limits.
\end{IEEEkeywords}

\section{Introduction}
Given a $m$-dimensional primary task to be performed by a robot with $n$ joints, with $n>m$ (redundancy), a standard method to prevent violation of joint/Cartesian inequalities during motion is to resort to some form of artificial potentials~\cite{khatib1986real}, pushing away from their limits the joints and the control points on the robot body~\cite{khatib2020task}. However, this method is highly parameter-dependent and may introduce oscillations when activating/deactivating the avoidance task in proximity of the bounds~\cite{khatib2020Irim}. To milden such undesired behavior, the null-space projection term or the activation function can be designed in an incremental way~\cite{mansard2009unified,simetti2016novel}. Nonetheless, selection of appropriate gains is still needed. Moreover, in case of multiple tasks, incorporating the avoidance behavior in the original Stack of Tasks (SoT) will assign different priorities to each single constraint~\cite{mansard2009unified,simetti2016novel,sentis2005synthesis,sentis2006whole}.

Other numerical approaches incorporate joint space and Cartesian motion limits as inequality constraints using parameter-free optimization, such as Quadratic Programming (QP)~\cite{escande2014hierarchical,hoffman2018multi}. However, these methods are computationally slower than analytical solutions and do not lead to realizable solutions when the original task(s) is not compatible with the set of inequality constraints. The SNS algorithm introduced in~\cite{flacco2015control} links QP to the SoT approach and overcomes these challenges. In the original paper, joint motion limits were considered as hard bounds (i.e., that cannot be relaxed in a least-square sense) and treated out of the SoT. On the other hand, Cartesian (avoidance) constraints were not handled as hard bounds. In~\cite{osorio2019physical}, an approach has been proposed to include joint and Cartesian inequality constraints in the SoT for torque-controlled manipulators. However, joint limits are always pre-assigned the highest priority over all other constraints. Moreover, all violated constraints are set to their limits at the saturation level, as opposed to what happens in~\cite{flacco2015control}. This is neither necessary nor optimal, and will often lead to high-frequency oscillations at the level of commands.

In this paper, we propose several improvements to the SNS algorithm at the velocity level introduced in~\cite{flacco2015control}. First, both joint and Cartesian inequality (box) constraints are treated as hard bounds. Second, all inequalities are assigned the same priority, i.e., enforced anyway independently of the {\em end-effector} (EE) task (or simply considered out of the SoT, in case of multiple tasks). Finally, the modifications are made so as to preserve the automatic optimal task scaling of the original SNS approach, which relaxes the primary task (keeping its geometric direction) only when no feasible solution would exist, while guaranteeing satisfaction of all inequality constraints. The resulting algorithm does not need parameter tuning and is faster than QP solvers.
\vspace{-5pt}

\section{Methodology}

\subsection{Generalised constraints}

Consider a robot with $n$ joints and $r$ generic Cartesian control points distributed on the robot body, each of dimension $d_i \in \{1,2,3\}$, $i=1,\dots, r$. We define an augmented vector
\begin{equation}
\av = \left(\!\begin{array}{ccccc}
\qv^T & \pv_{cp,1}^T & \pv_{cp,2}^T & \dots & \pv_{cp,r}^T\end{array}\!\right)^T,
\label{eq:a}
\end{equation}
where $\qv \in \mathbb{R}^n$ denotes the joint variables and $\pv_{cp,i}\in \mathbb{R}^{d_i}$ is the position of the $i$-th control point, $i=1,\dots,r$. The joint variables as well as the Cartesian control points will have some desired motion restrictions. Accordingly, we define the augmented matrix 
\begin{equation}
\Am = \left(\!\begin{array}{ccccc}
		\IIm & \Jm_{cp,1}^T & \Jm_{cp,2}^T & \dots & \Jm_{cp,r}^T
\end{array}\!\right)^T,
\label{eq:augment_mat}
\end{equation}
where $\IIm \in \mathbb{R}^{n \times n}$ is the identity matrix and $\Jm_{cp,i} \in \mathbb{R}^{d_i \times n}$ is the Jacobian matrix of the $i$-th control point.
Define the position and velocity limits for each joint, $j=1,\dots,n$, as
\begin{equation}
Q_j^{min} \leq q_j \leq Q_j^{max}, 
\qquad 
V_j^{min} \leq \dot{q}_j \leq V_j^{max},
\label{eq.4.1j}
\end{equation}
and the limits for each control point, $i=1,\dots,r$, as
\begin{equation}
\Pm_{cp,i}^{min} \leq \pv_{cp,i} \leq \Pm_{cp,i}^{max}, 
\qquad
\Vm_{cp,i}^{min} \leq \dpv_{cp,i} \leq \Vm_{cp,i}^{max}.
\label{eq.4.1}
\end{equation}
At a generic time instant, the box constraints for the velocity of each component of~(\ref{eq:a}) are given by
\begin{equation}
\begin{array}{rcl}
    \dot{Q}_{min,j} &\!\!\!\!=\!\!\!\!& \displaystyle \max \left\{ \frac{Q_j^{min}-q_j}{T},V_j^{min} \right\},\\[12pt]
    \dot{Q}_{max,j} &\!\!\!\!=\!\!\!\!& \displaystyle \min \left\{ \frac{Q_j^{max}-q_j}{T},V_j^{max} \right\}, 
\end{array}
\label{eq:cons1}
\end{equation}
and
\begin{equation}
\begin{array}{rcl}
    \dot{\Pm}_{cp,i}^{min}&\!\!\!\!=\!\!\!\!& \displaystyle \max \left\{ \frac{\Pm_{cp,i}^{min}-\pv_{cp,i}}{T},\Vm_{cp,i}^{min} \right\}, \\[12pt]
    \dot{\Pm}_{cp,i}^{max}&\!\!\!\!=\!\!\!\!&\displaystyle \min \left\{ \frac{\Pm_{cp,i}^{max}-\pv_{cp,i}}{T},\Vm_{cp,i}^{max} \right\},
\end{array}
\label{eq:cons2}
\end{equation}
where $T$ is the sampling time. Accordingly, the generalised inequality constraints can be written as the augmentation of joint and Cartesian bounds in~(\ref{eq:cons1}) and~(\ref{eq:cons2}) as
\begin{equation}
\begin{array}{rcl}
\Bm_{min} &\!\!\!\!=\!\!\!\!& \left(\!\begin{array}{cccccc}
\dot{Q}_{min,1}, & \!\!\dots\!\! & \dot{Q}_{min,n}, &\!
\dot{\Pm}_{cp,1}^{min}, & \!\!\dots\!\! & \dot{\Pm}_{cp,r}^{min}
\end{array}\!\right)^T\!, \\[4pt]
\Bm_{max} &\!\!\!\!=\!\!\!\!& \left(\!\begin{array}{cccccc}
\dot{Q}_{max,1}, & \!\!\dots\!\! & \dot{Q}_{max,n}, &\!
\dot{\Pm}_{cp,1}^{max}, & \!\!\dots\!\! & \dot{\Pm}_{cp,r}^{max}
\end{array}\!\right)^T\!.
\end{array}
\label{eq:Gbound}
\end{equation}

\subsection{Modified SNS algorithm}

Consider a single EE velocity task $\dxv_d \in \mathbb{R}^m$, with $n>m$, and its Jacobian matrix $\Jm \in \mathbb{R}^{m \times n}$, to be achieved under the generalised hard constraints in~(\ref{eq:Gbound}). The pseudo-code of the modified SNS method is presented in \Cref{basic:vel:single}. If the Cartesian inequality limits in~(\ref{eq.4.1}) are discarded, the augmented matrix~(\ref{eq:augment_mat}) becomes $\Am = \IIm$ and it is easy to show that \Cref{basic:vel:single} simplifies to the original SNS algorithm in~\cite{flacco2015control}.
Note that at line~15, the most critical constraint corresponds to the smallest scaling factor $s_k$ over all constraints. Also, at line~19, when there is no way to perform the desired task under the hard inequality constrains, we apply an optimal task scaling factor as computed by~\Cref{scaling_factor}, similar to~\cite{flacco2015control}. 

\begin{algorithm}[t]
	\caption{SNS with generalised inequality constraints}
	\begin{algorithmic}[2]
		\State$\dqv_N \gets \mathbf{0},\; s^{\ast} \gets 0,\; \Pm \gets \IIm, \; \Am_{lim} \gets \text{null},\; \dot{\av}_N \gets \text{null}$
		\Repeat
		\State $\text{limits\_violated} \gets \text{FALSE}$
		
		\State $ \dqv \gets \dqv_N + \left( \Jm\,\Pm \right)^{\#}  \left( \dxv - \Jm\,\dqv_N \right)$
		\State $ \dot{\av} \gets \Am\,\dqv $
		\If{$ \exists \,h \in \, \left[1:n+\Sigma_1^r d_i\right]: \,  (\dot{a}_h < {{b}}_{min,h}) \, \lor \,  (\dot{a}_h > {{b}}_{max,h})  $}
		\State $\text{limits\_violated} \gets \text{TRUE}$
		\State $ \boldsymbol{\alpha} \gets \Am \left( \Jm\,\Pm \right)^{\#}\dxv$
		\State $ \boldsymbol{\beta} \gets \dot{\av} - \boldsymbol{\alpha}$
		\State $ \text{getTaskScalingFactor}(\boldsymbol{\alpha},\boldsymbol{\beta}) $
		\If{$ \{\text{task scaling factor}\} > s^\ast$}
		\State $ s^\ast \gets  \{\text{task scaling factor}\} $
		\State $ \dqv_N^\ast \gets \dqv_N,\;\Pm^\ast \gets  \Pm$
		\EndIf
		\State $ k \gets \{\text{the most critical constraint}\} $
		\State $ \Am_{lim} \gets \text{concatenate}(\Am_{lim},\Am_k) $
		\State $ \dot{a}_{N} \gets \begin{cases}
			\text{concatenate}(\dot{a}_{N},{{b}}_{max,k})\;\;\;\;\;\; \text{if}\;\;(\dot{a}_h > {{b}}_{max,k})\\
			\text{concatenate}(\dot{a}_{N},{{b}}_{min,k})\;\;\;\;\;\; \text{if}\;\;(\dot{a}_h < {{b}}_{min,k})\\\end{cases} $
		\State $ \Pm \gets  \IIm - \left( \Am_{lim} \right)^{\#}\left(\Am_{lim}\right)$
		\If{$ \text{rank}(\Jm \Pm) < m$}
		\State $ \dqv \gets \dqv_N^\ast + \left( \Jm\,\Pm^\ast \right)^{\#}  \left( s^\ast\dxv - \Jm\,\dqv_N^\ast \right)  $
		\State $\text{limits\_violated} \gets \text{FALSE} $ 
		\EndIf
		\EndIf
		\State $ \dqv_N \gets \left( \Am_{lim} \right)^{\#} \, \dot{a}_{N} $
		\Until{$\text{limits\_violated} = \text{TRUE}$}
		\State $ \dqv_{SNS} \gets \dqv$
	\end{algorithmic}
\label{basic:vel:single}
\end{algorithm}

\begin{algorithm}[t]
	\caption{Optimal task scaling factor}
	\begin{algorithmic}[2]
    \Function{getTaskScalingFactor}{$\boldsymbol{\alpha},\boldsymbol{\beta}$}
        \For {$h \leftarrow 1: n+\Sigma_1^r d_i$}
            \State $L_h \gets b_{min,h}-\beta_h$
            \State $U_h \gets b_{max,h}-\beta_h$
                \If{$ {\alpha}_{h} < 0 \, \land \,  L_i < 0 $}
                    \If{$ \alpha_h < L_h $}
                    \State $s_h \gets L_h^{}/\alpha_h$
                    \Else
                    \State $s_h \gets 1 $
                    \EndIf
                \ElsIf{$\alpha_h>0 \, \land \, U_h > 0$}
                    \If{$ \alpha_h > U_h $}
                    \State $s_h \gets U_h^{}/\alpha_h$
                    \Else
                    \State $s_h \gets 1 $
                    \EndIf
                \Else
                    \State $s_h \gets 0$
                \EndIf
        \EndFor\\
    \Return $s$
    \EndFunction
	\end{algorithmic}
\label{scaling_factor}
\end{algorithm}

\section{Results}

\begin{figure}[t]
	\centering
	\includegraphics[width=\linewidth]{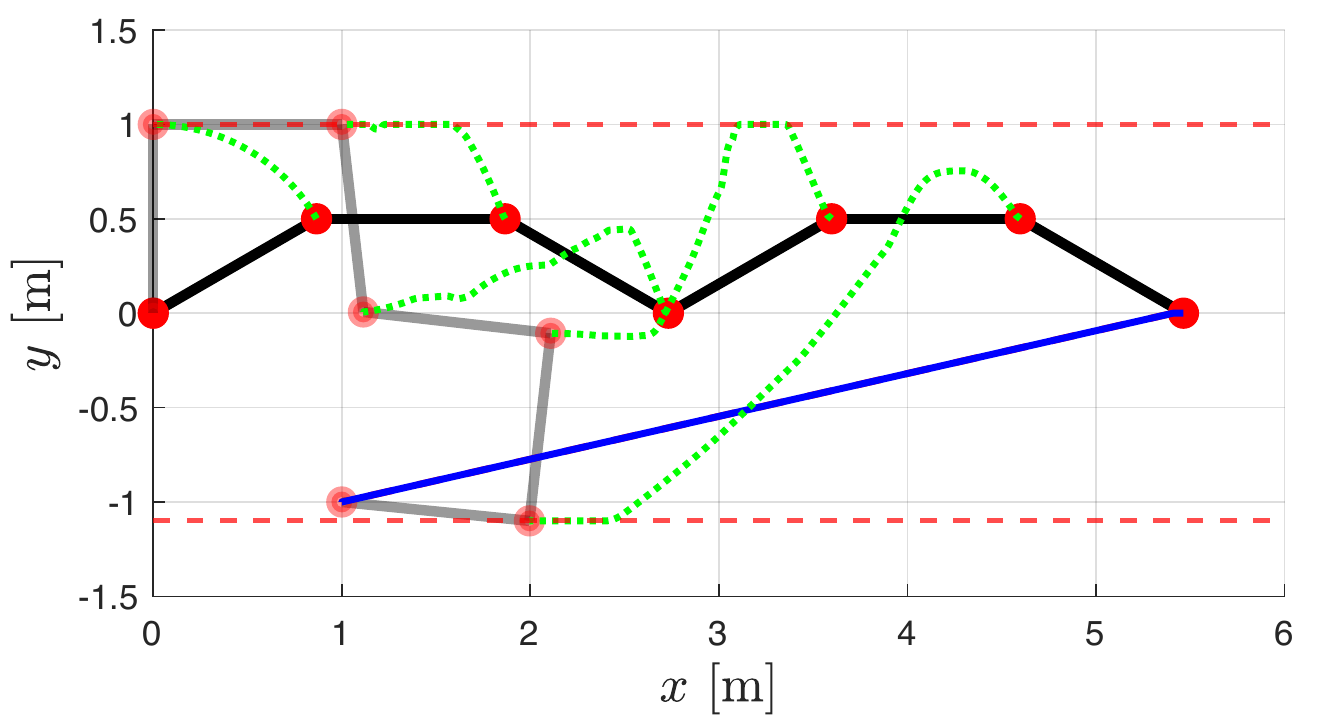}  
	\caption{Initial (black) and final (gray) configurations of the 6R planar arm. The red circles represent the robot joints (and the EE tip). The desired EE path is the blue line, to be traced from right to left. The dashed red lines are the Cartesian position limits. The dashed green lines show the path of control points during task execution.}
	\label{fig:motion}
\vspace{-5pt}
\end{figure}

\begin{figure}[h]
	\centering
	\includegraphics[width=\linewidth]{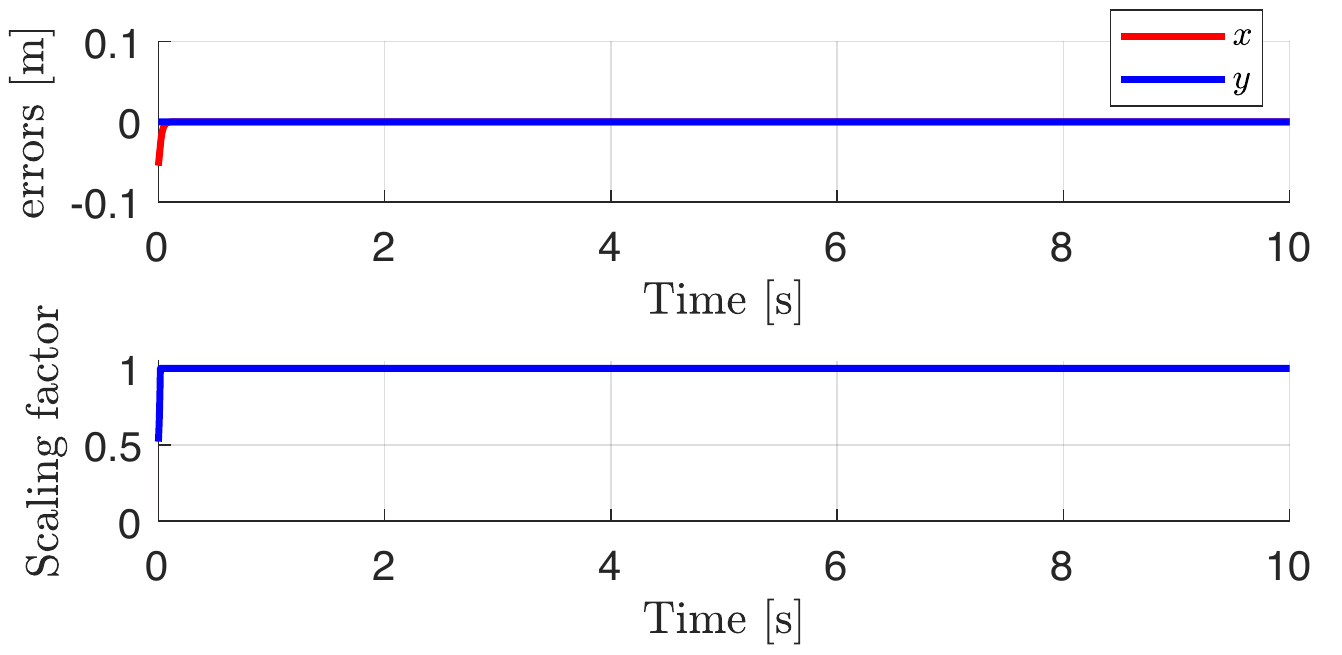}
	\caption{EE positional errors and related task scaling factor.}
	\label{fig:error}
\vspace{-10pt}
\end{figure}

\begin{figure}[h]	
	\centering
	\begin{subfigure}{0.45\textwidth}
		\centering
		\includegraphics[width=\linewidth]{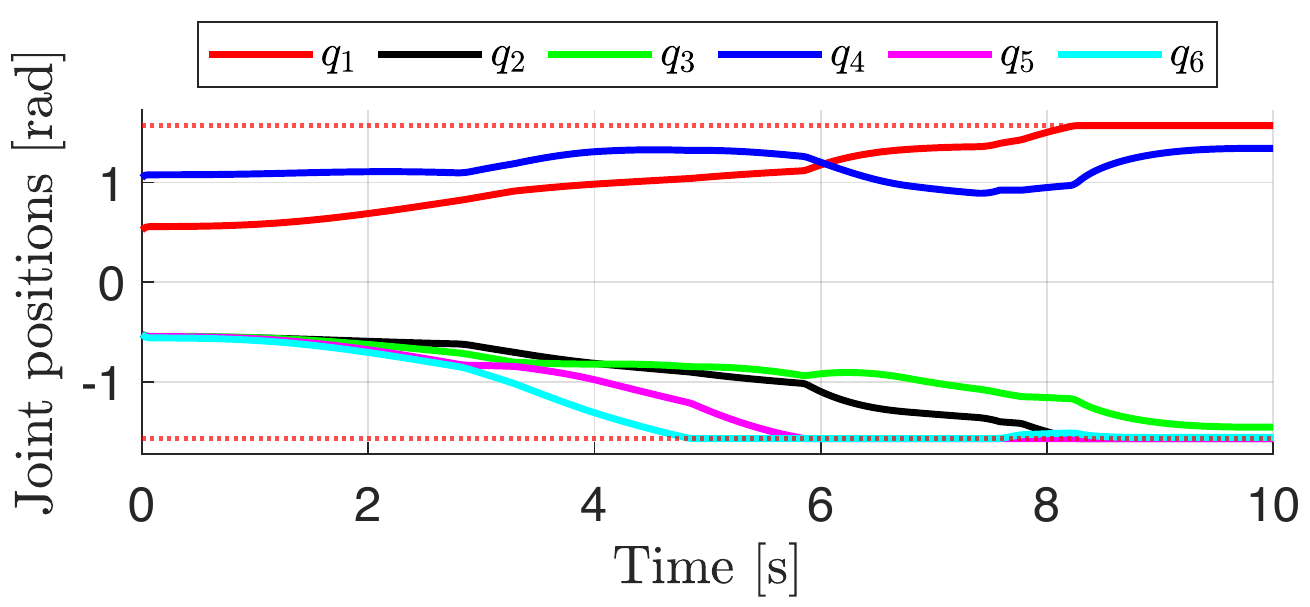}
	\end{subfigure}
	\begin{subfigure}{0.45\textwidth}
		\centering
		\includegraphics[width=\linewidth]{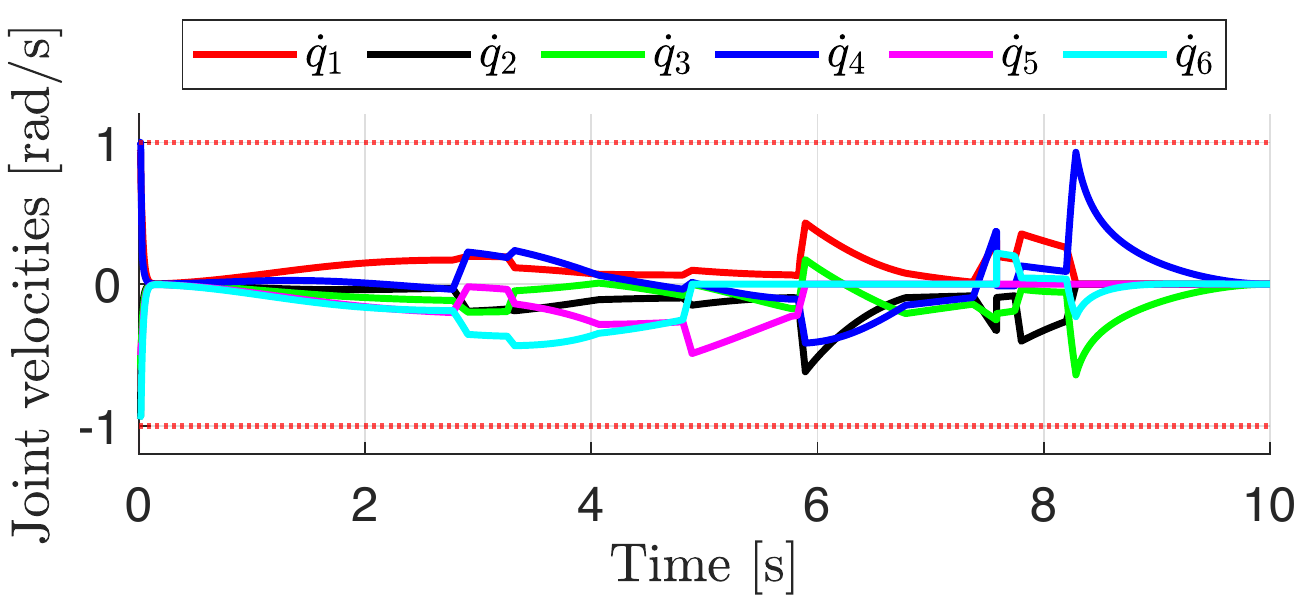} 
	\end{subfigure}
	\caption{Evolution of the joints during task execution. The dotted red lines are the bounds on the joint motion.}
	\label{fig:joints}
\vspace{-5pt}
\end{figure}

\begin{figure}[h]		
\centering
	\begin{subfigure}{0.45\textwidth}
		\centering
		\includegraphics[width=\linewidth]{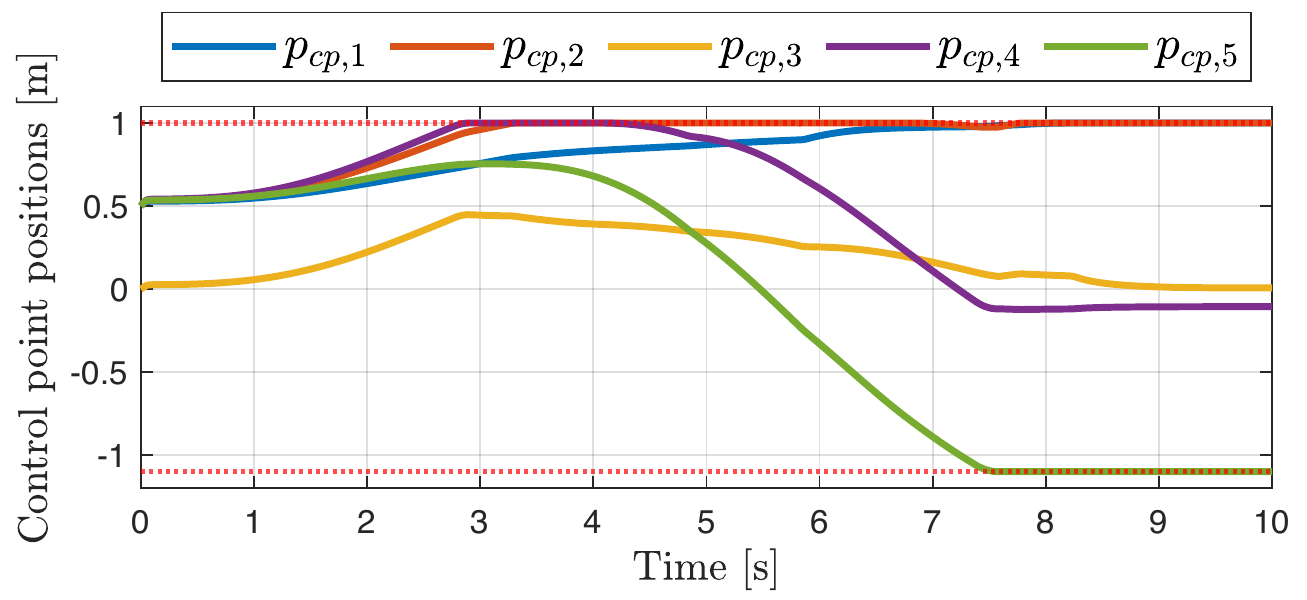}  
		\end{subfigure}	
	\begin{subfigure}{0.45\textwidth}
		\centering
		\includegraphics[width=\linewidth]{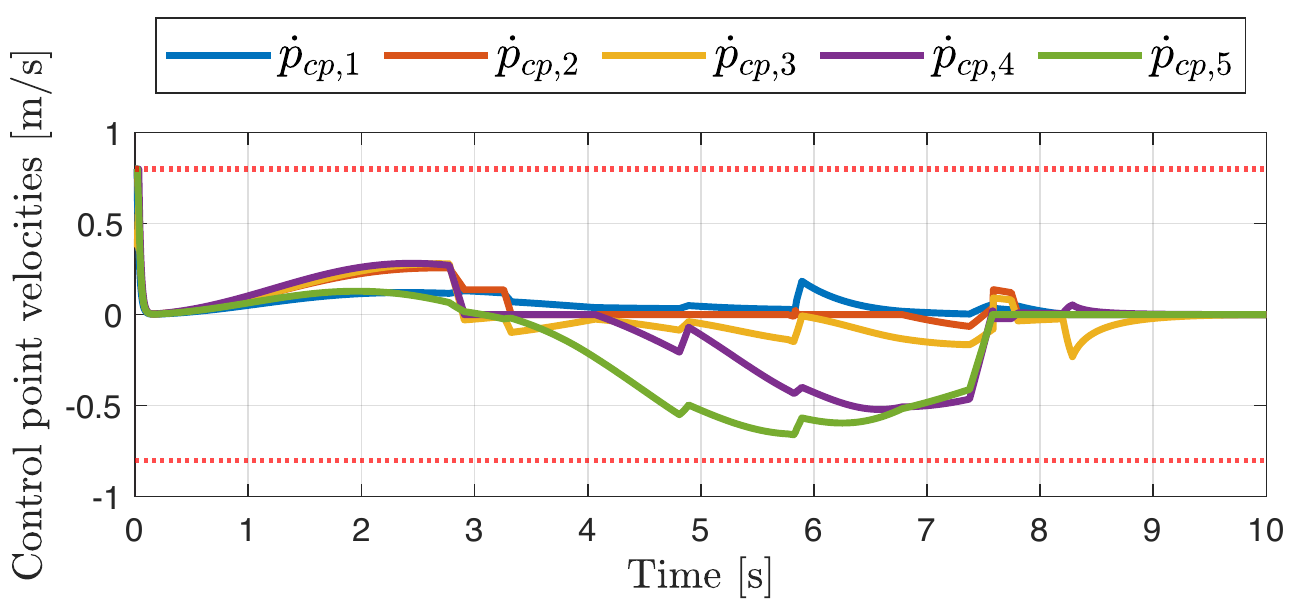}  
		\end{subfigure}
	\caption{Evolution of the control points along the $y$-direction. The dotted red lines are the Cartesian bounds on the motion of the control points.}
	\label{fig:cartesian}
\vspace{-10pt}
\end{figure}

Verification for the proposed algorithm is done through MATLAB simulations, using a 6R planar robot arm ($n=6$) and considering joint and Cartesian inequality constraints. The EE is required to track a 2D linear path ($m=2$), see Fig.~\ref{fig:motion}. Therefore, the primary EE velocity task is defined as 
\begin{equation}
\dxv = \dxv_d + \Km_p(\xv_d - \xv_{ee}),
\end{equation}
with the control gain matrix $\Km_p = \mbox{diag}\{50, 50\}$ and the EE position $\xv_{ee}$ computed by the direct kinematics. The sampling time is $T = 1$~[ms] and the initial robot configuration (in [rad]) is chosen as
\begin{equation}
\qv_0 = \left(\!\begin{array}{cccccc} \displaystyle \frac{\pi}{6} & \displaystyle -\frac{\pi}{6} &\displaystyle -\frac{\pi}{6} & \displaystyle\frac{\pi}{3}& \displaystyle-\frac{\pi}{6}& \displaystyle -\frac{\pi}{6} \end{array}\!\right)^T.
\end{equation}
The limits~(\ref{eq.4.1j}) are equal and symmetric for all joints:
\begin{equation}
{Q}_j^{max} = -{Q}_j^{min} = \frac{\pi}{2} \, \mbox{[rad]}, \quad
{V}_j^{max} = -{V}_j^{min} = 1 \, \mbox{[rad/s]}.
\end{equation}
We consider $r=5$ control points (each with $d_i=1$) along the robot body, located at the joints $j=2,\dots,6$. The corresponding limits~(\ref{eq.4.1}) are equal for all points, and imposed only over the $y$-direction:
\begin{equation}
\begin{array}{l}
P_{cp,i}^{max,y} = 1 \, \mbox{[m]}, \quad P_{cp,i}^{min,y} = -1.1 \, \mbox{[m]}, 
\\[4pt]
V_{cp,i}^{max,y} = -V_{cp,i}^{min,y} =0.8 \,\mbox{[m/s]}.
\end{array}
\end{equation}
The EE starts the motion very close to the desired path. As shown in Fig.~\ref{fig:error}, the positional error converges immediately and remains zero along the whole task. Accordingly, the task scaling is active ($s < 1$) only for few milliseconds at beginning, to comply with the saturated joint and Cartesian velocity limits due to the initial error recovery ---see Figs.~\ref{fig:joints} and~\ref{fig:cartesian}. Later, the robot is able to perform the complete task perfectly while satisfying all inequality constraints (many of them in saturation). The few discontinuities in the commanded joint velocity in~Fig.~\ref{fig:joints} can be addressed by extending the algorithm to the acceleration level and including suitable joint acceleration limits in the set of constraints. 

\section{Conclusion}

We have proposed major enhancements to the basic SNS algorithm at the velocity level for redundant robots. Cartesian inequality constraints are included and treated as hard limits, while preserving all the nice features of the original method. The modified algorithm can be extended to include multiple tasks having different priorities. Moreover, it can be implemented also at the acceleration level, which is beneficial for involving dynamic properties in the resolution of redundancy and is suitable for torque-controlled systems.

\bibliographystyle{ieeetr}
\bibliography{refs}
\end{document}